\documentclass[sigconf]{acmart}

\copyrightyear{2024}
\acmYear{2024}
\setcopyright{acmlicensed}\acmConference[WWW '24]{Proceedings of the ACM Web Conference 2024}{May 13--17, 2024}{Singapore, Singapore}
\acmBooktitle{Proceedings of the ACM Web Conference 2024 (WWW '24), May 13--17, 2024, Singapore, Singapore}
\acmDOI{10.1145/3589334.3645385}
\acmISBN{979-8-4007-0171-9/24/05}
\usepackage{appendix}
\usepackage{multirow}
\usepackage{subcaption}
\usepackage{graphicx}
\usepackage{array}   
\usepackage{svg}
\usepackage{balance}
\usepackage{multicol}
\newcommand{\haorui}[1]{\textcolor{black}{#1}} % red
\newcommand{\camera}[1]{\textcolor{black}{#1}} % red
\newcommand{\todo}[1]{\textcolor{black}{#1}} % red
\newcolumntype{C}[1]{>{\centering\arraybackslash}p{#1}}

\begin{document}

\title{MCFEND: A Multi-source Benchmark Dataset for Chinese Fake News Detection}
\thanks{Haorui He and Dacheng Wen are also with Department of Interactive Media, Hong Kong Baptist University, Hong Kong, China. This work was done while Haorui He, Jin Bai, and Dacheng Wen were under the supervision of Yupeng Li. Haorui He is the corresponding author.}
\thanks{The MCFEND Dataset is available at \todo{\url{https://github.com/TrustworthyComp}}.}
\thanks{This is the author's version of the work. It is posted here for your personal use. Not for redistribution. The definitive Version of Record will be published in the Proceedings of the ACM Web Conference 2024.}

\author{Yupeng Li}
\affiliation{%
  \department{Department of Interactive Media}
  \institution{Hong Kong Baptist University}
  \city{Hong Kong}
  \country{China}
}
\email{ivanypli@gmail.com}

\author{Haorui He}
\affiliation{%
  \department{School of Data Science}
  \institution{The Chinese University of Hong Kong, Shenzhen}
\city{Shenzhen}
  \country{China}
}
\email{hehaorui11@gmail.com}

\author{Jin Bai}
\affiliation{
  \department{Department of Computer Science}
  \institution{Beijing Normal University-Hong Kong Baptist University United International College}
  \city{Zhuhai}
  \country{China}
}
\email{jinbai@uic.edu.cn}

\author{Dacheng Wen}
\affiliation{%
  \department{Department of Computer Science}
  \institution{The University of Hong Kong}
    \city{Hong Kong}
  \country{China}
}
\email{wdacheng@connect.hku.hk}
\renewcommand{\shortauthors}{Yupeng Li, Haorui He, Jin Bai, and Dacheng Wen}

\begin{abstract}
The prevalence of fake news across various online sources has had a significant influence on the public. Existing Chinese fake news detection datasets are limited to news sourced solely from Weibo. However, fake news originating from multiple sources exhibits diversity in various aspects, including its content and social context. Methods trained on purely one single news source can hardly be applicable to real-world scenarios. Our pilot experiment demonstrates that the F1 score of the state-of-the-art method that learns from a large Chinese fake news detection dataset, \textit{Weibo-21}, drops significantly from 0.943 to 0.470 when the test data is changed to multi-source news data, failing to identify more than one-third of the multi-source fake news. To address this limitation, we constructed the first multi-source benchmark dataset for Chinese fake news detection, termed \texttt{MCFEND}, which is composed of news we collected from diverse sources such as social platforms, messaging apps, and traditional online news outlets. Notably, such news has been fact-checked by 14 authoritative fact-checking agencies worldwide. In addition, various existing Chinese fake news detection methods are thoroughly evaluated on our proposed dataset in \emph{cross-source}, \emph{multi-source}, and \emph{unseen source} ways. \texttt{MCFEND}, as a benchmark dataset, aims to advance Chinese fake news detection approaches in real-world scenarios.
\end{abstract}

\begin{CCSXML}
<ccs2012>
   <concept>
       <concept_id>10002951.10003227.10003351</concept_id>
       <concept_desc>Information systems~Data mining</concept_desc>
       <concept_significance>500</concept_significance>
       </concept>
    <concept>
       <concept_id>10002944.10011123.10011130</concept_id>
       <concept_desc>General and reference~Evaluation</concept_desc>
       <concept_significance>500</concept_significance>
       </concept>
   <concept>
       <concept_id>10010147.10010178.10010179</concept_id>
       <concept_desc>Computing methodologies~Natural language processing</concept_desc>
       <concept_significance>500</concept_significance>
       </concept>
 </ccs2012>
\end{CCSXML}

\ccsdesc[500]{Information systems~Data mining}
\ccsdesc[500]{General and reference~Evaluation}
\ccsdesc[500]{Computing methodologies~Natural language processing}

\keywords{Chinese Fake News Detection, Multi-source Benchmark Dataset, Cross-source Evaluation, Multi-source Evaluation}

\maketitle

\section{Introduction}
\label{sec:intro}
It has been prevalent for the public to consume news through various online sources, such as social platforms and news websites. Such sources are efficient media for spreading fake news. For instance, the latest Weibo annual report on fake news~\cite{weibo-report} revealed that Weibo's official fact-checking agency identified 82,274 pieces of fake news in 2022. Given the devastating consequences of fake news on both individuals and society, fake news detection has become an urgent and essential task that needs to be addressed~\cite{support5,survey,WEIBO21, EANN, fake1, fake2, fake3}. To this end, Chinese fake news detection datasets have been constructed for the development of Chinese fake news detection~\cite{survey, WEIBO16, Media-Weibo, BERT-EMO, WEIBO21, CHECKED, MR2}.

\begin{figure}[h]
    \centering
    \includegraphics[width=\linewidth]{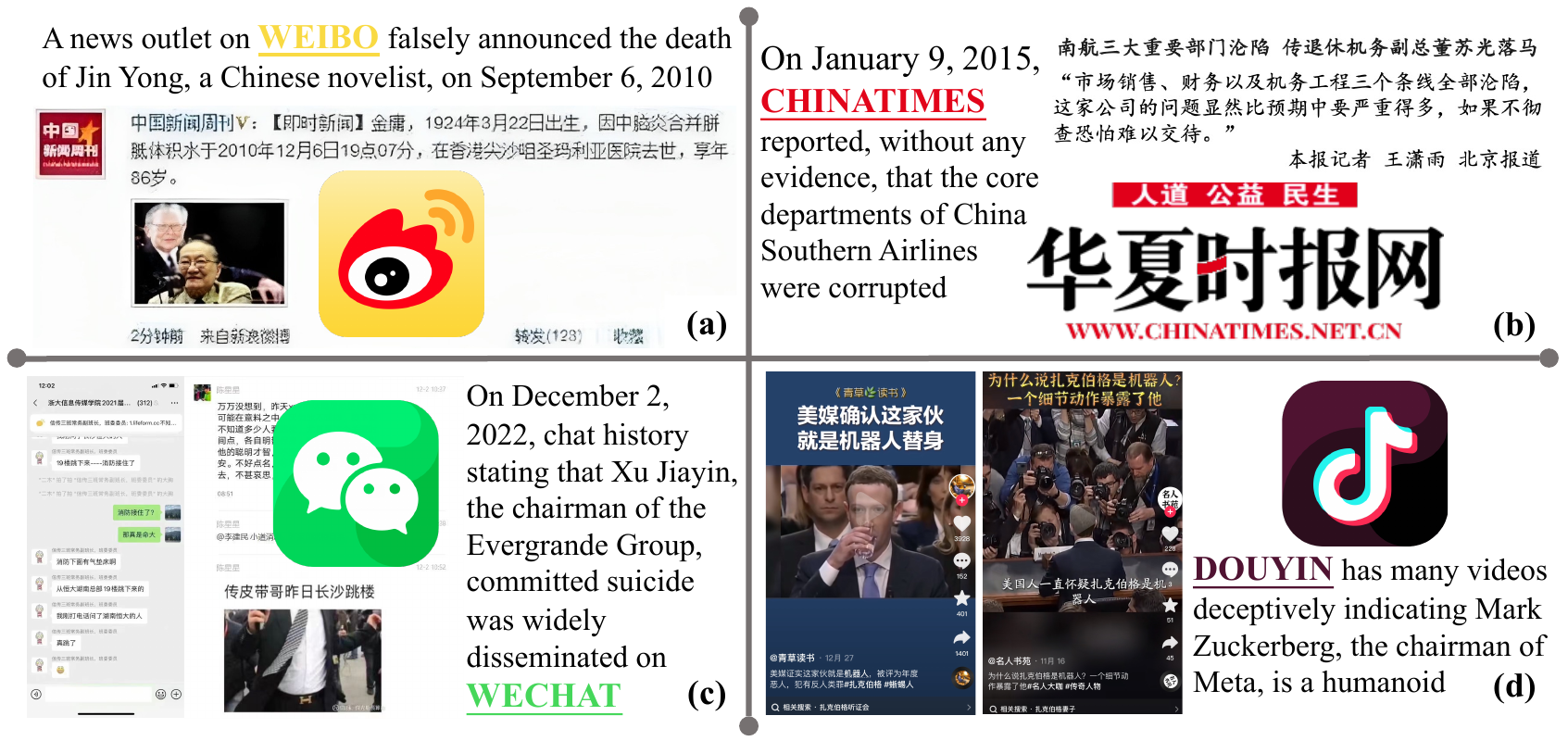}
    \caption{An example of four pieces of fake news from four different Chinese news sources, including Weibo (a popular social platform), China Times (an online news outlet), Wechat (a messaging app), and Douyin (a social platform). Each piece of fake news showcases different characteristics across various aspects, such as content, topics, publishing methods, linguistic styles, etc.}
    \label{fig:example}
\end{figure}

The existing Chinese fake news detection datasets are limited to one single source, Weibo, where both true and fake news are collected. However, in the real world, news emerges from multiple sources, such as social platforms, messaging apps, and traditional online news outlets, etc.~\cite{survey,survey2,survey3,stance1,stance2}. News, in particular, fake news, from different sources is characterized by diverse dimensions, such as news content, topics, publishing methods, and the utilization of sophisticated linguistic styles intended to mimic real news~\cite{support1,support2,support3, support4,support5}. For example, Fig.~\ref{fig:example} shows four instances of fake news, each sourced from a distinct news source, exemplifying different news characteristics. Existing Weibo-based Chinese fake news detection datasets that fail to capture the above data diversity can impede the effectiveness of machine learning (ML) based fake news detection in practice, including but not limited to the robustness to intricately crafted fake news and the generalization to fake news from other sources~\cite{Trustworthy, support4, support5, survey, WEIBO21}.

\textbf{Pilot Experiment}. To verify such limitations, we conducted evaluations on 817 pieces of fake news we collected. They were verified between Jan. 2015 and Mar. 2023, from the China Internet Joint Rumor Refuting Platform\footnote{\url{https://www.piyao.org.cn/}}, a government-backed fact-checking agency supported by authoritative experts and various government departments. The agency covers fake news originating from a wide variety of sources, including but not limited to, Douyin, Wechat, TouTiao, Zhihu, Weibo, etc.\footnote{
\camera{The websites for the mentioned news sources are as follows: 
Douyin: \url{https://www.douyin.com/};
Wechat: \url{https://www.wechat.com/}; 
TouTiao: \url{https://www.toutiao.com/}; 
Zhihu: \url{https://www.zhihu.com/}; and Weibo: \url{https://m.weibo.cn/}.}
} 
We trained the state-of-the-art fake news detection model BERT-EMO~\cite{BERT-EMO} on the \textit{Weibo-21} dataset~\cite{WEIBO21}. 
The model demonstrated strong performance with F1 scores of 0.943 on the \textit{Weibo-21}, 0.932 on the \textit{Weibo-20}~\cite{BERT-EMO} dataset, and 0.908 on the \textit{Weibo-16} dataset~\cite{WEIBO16}, respectively.
Nevertheless, when we used the same model to detect fake news collected from all the diverse sources on the platform,  failed to identify 35.34\% of fake news. The macro F1 score dropped to 0.470, a decline of 52.03\%. 
The following may account for such experimental results. 
First, during the training phase, ML models lack exposure to a diverse range of data from various sources, which leads to an overfitting of specific characteristics of Weibo fake news. This limits their capability to generalize and effectively identify fake news from different sources. 
Further, during the testing phase, these models are evaluated using Weibo data exclusively, overlooking a comprehensive assessment of their performance across different news sources. 
When faced with real-world fake news that emerges from multiple sources, the applicability of models trained and tested on existing datasets is questionable. Our analytical results validate the limitations of current Chinese fake news detection datasets. Therefore, it is imperative to construct a comprehensive dataset that consists of (real and fake) from diverse sources.

To bridge this gap, we constructed the first \underline{M}ulti-source benchmark dataset for \underline{C}hinese \underline{F}ak\underline{E} \underline{N}ews \underline{D}etection (\texttt{MCFEND}), which contains 23,789 pieces of authoritatively verified Chinese news from \emph{14 fact-checking agencies covering numerous news sources.}\footnote{The Weibo Community Management Center, Weibo's official fact-checking agency, exclusively examines news sourced from Weibo, whereas other fact-checking agencies cover various news sources. Note that the inclusion of additional fact-checking agencies can potentially enhance the variety of news sources considered. Please refer to Table~\ref{tab:source} for the full list of included fact-checking agencies.} These fact-checking agencies are further divided into three distinct groups. The first group encompasses nine Chinese fact-checking agencies. That means the collected news has been verified by experts as active and authoritative. The second group corresponds to three existing annotated English fake news detection datasets. Specifically, for an English news piece paired with its corresponding authenticity label from an existing English fake news dataset, we employ the state-of-the-art cross-lingual identical news retrieval system to collect its Chinese equivalent while retaining its original label. The third group consists of Weibo's official fact-checking agency exclusively, i.e., the Weibo Community Management Center. For this group of news, we utilized news data from the \textit{Weibo-21} dataset~\cite{WEIBO21}. Furthermore, we conducted comprehensive evaluations on six established baseline models for Chinese fake news detection, including state-of-the-art methods, under cross-source and multi-source scenarios on our \texttt{MCFEND} dataset. The experimental results characterize the challenge of accurately spotting fake news from different sources that the dataset presents.

Our contributions are summarized as follows:
First, we constructed the initial multi-source Chinese fake news detection dataset \texttt{MCFEND}, which comprises multi-modal content and social context of 23,789 real-world Chinese news pieces collected from 14 authoritative fact-checking agencies in three distinct groups. To the best of our knowledge, \texttt{MCFEND} is the largest open-sourced Chinese fake news detection dataset, being at least 2.63 times larger than any existing ones. The dataset aims to benchmark the evaluations of Chinese fake news detection methods in real-world scenarios, where news originates from diverse sources, and to encourage further research in this field.
Second, we conducted comprehensive cross-source, multi-source, and unseen source evaluations on six established baseline models for Chinese fake news detection, including state-of-the-art methods. Our experimental results reveal that the models trained on existing datasets are not applicable in real-world scenarios. Incorporating multi-source data is necessary, which can enhance the models' robustness substantially.

\section{Preliminaries and Related work}\label{sec:relate}
Fake news detection, also referred to as false news detection, and related to information credibility evaluation, is commonly defined as a binary classification task~\cite{survey, support4, survey2, survey3}. 
In this context, we define the output space as $\mathcal{Y} = \{0, 1\}$ to classify news items as real ($0$) or fake ($1$). The input space $\mathcal{X}$ includes multidimensional data, comprising both the content of the news and its social context. 
Let $\mathcal{D} = \{ (x_i, y_i) \}_{i=1}^n$ represent a collection of $n$ news pieces, each associated with a label indicating its authenticity. Here, $x_i \in \mathcal{X}$ and $y_i \in \mathcal{Y}$.
The goal is to design fake news detection models capable of learning a function $\phi: \mathcal{X} \rightarrow \mathcal{Y}$, such that for any given news item $x_i$, $\phi(x_i)$ predicts its label $y_i$ with high accuracy.

Numerous datasets have been constructed to address fake news detection. Representative English fake news detection datasets, such as \textit{BuzzFace}~\cite{BuzzFace}, \textit{LIAR}~\cite{LIAR}, \textit{FakeNewsNet}~\cite{FAKENEWSNET}, \textit{PHEME}~\cite{PHEME}, \textit{KaggleFakeNews}~\cite{kagglefakenews}, \textit{FakeNewsCorpus}~\cite{support3}, and \textit{FakeHealth}~\cite{FAKEHEALTH}, were constructed on English news from social platforms like Twitter and Facebook, as well as fact-checking websites such as BuzzFeed, PolitiFact, and NewsGuard.\footnote{The websites for these social platforms are as follows: Twitter: \url{https://www.twitter.com/}; Facebook: \url{https://www.facebook.com/}; BuzzFeed: \url{https://www.buzzfeed.com/}; PolitiFact: \url{https://www.politifact.com/}; and NewsGuard: \url{https://www.newsguardtech.com}.} A few Chinese fake news detection datasets have also been proposed. For instance, Ma et al. introduced the \textit{Weibo-16} dataset~\cite{WEIBO16}, collected from the Chinese social platform Weibo. This dataset contains verified fake news sourced from the Weibo Community Management Center\footnote{\url{https://service.account.weibo.com}}, an official fact-checking agency for posts on Weibo. Real news was collected from regular posts that were not categorized as fake. While \textit{Weibo-16} focuses exclusively on textual data, Jin et al.~\cite{Media-Weibo} later introduced \textit{Media-Weibo}, the first multi-modal dataset for detecting Chinese fake news. \textit{Media-Weibo} includes textual content, user profiles, and supplementary images for each post. Zhang et al.~\cite{BERT-EMO} then extended \textit{Media-Weibo} dataset to \textit{Weibo-20} dataset by adding 850 real news pieces authenticated by NewsVerify\footnote{\url{https://www.newsverify.com/}}, a fact-checking website dedicated to verifying posts on Weibo, from April 2014 to November 2018, and 1,806 fake news pieces that were officially verified by the Weibo Community Management Center within the same timeframe. Using a similar approach, Yang et al.~\cite{CHECKED} constructed \textit{CHECKED} dataset, aiming at detecting COVID-19-related fake news on Weibo. Additionally, Nan et al.~\cite{WEIBO21} proposed the \textit{Weibo-21} dataset, the first multi-domain Chinese fake news detection dataset. \textit{Weibo-21} contains both fake and real news pieces collected from Weibo spanning from December 2014 to March 2021, covering nine different domains, such as Science, Military, and Education. Most recently, Hu et al.~\cite{MR2} constructed the multi-modal retrieval augmented dataset \textit{MR2}. This dataset consists of two subsets from Weibo and Twitter, respectively, covering news with images and texts, and provides evidence retrieved from the Internet for both modalities.

\begin{table}[h]
\caption{Summary of Chinese fake news detection datasets. Please note that the \textit{MR2} dataset comprises two subsets: one from Weibo (Chinese) and one from Twitter (English). The statistics presented in this table specifically pertain to the Weibo (Chinese) subset.}
\label{table:datasets}
\begin{tabular}{cccccc}
\hline
\multirow{3}{*}{\textbf{Dataset}} &
\multirow{3}{*}{\textbf{\begin{tabular}[c]{@{}c@{}}\#News\end{tabular}}} &
\multicolumn{3}{c}{\textbf{Feature}} &
\multirow{3}{*}{\textbf{\begin{tabular}[c]{@{}c@{}}News\\ Source\end{tabular}}} \\ \cline{3-5}
 & & \textit{Text}    & \textit{Image}   & \textit{\begin{tabular}[c]{@{}c@{}}Social\\ Context\end{tabular}} &       \\ \hline
\textit{Weibo-16}    
& 5,656 & \checkmark &  & \checkmark & Weibo \\
\textit{Weibo-Media} 
& 5,802 & \checkmark & \checkmark & \checkmark & Weibo \\
\textit{Weibo-20}    
& 6,362  & \checkmark & \checkmark & \checkmark & Weibo \\
\textit{Weibo-21}    
& 9,128 & \checkmark & \checkmark & \checkmark & Weibo \\
\textit{MR2}         
& 6,976 & \checkmark & \checkmark & \checkmark & Weibo \\ 
\hline
\texttt{MCFEND}      
&23,789 & \checkmark & \checkmark & \checkmark & \begin{tabular}[c]{@{}c@{}}Multiple\\ Sources\end{tabular} \\ \hline
\end{tabular}
\end{table}

Existing datasets for Chinese fake news detection rely heavily on Weibo. Different from them, we constructed the pioneering multi-source Chinese fake news detection dataset, termed \texttt{MCFEND}, which contains 23,789 real-world Chinese news pieces collected from multiple sources across three distinct categories. Table~\ref{table:datasets} compares the Chinese fake news detection datasets.

\section{MCFEND Dataset}\label{sec:dataset}
In this section, we introduce our Chinese multi-source fake news detection dataset, \texttt{MCFEND}. Additionally, we perform data analysis to investigate into the differences between various sources.
 
\subsection{Overview}
The \texttt{MCFEND} dataset contains news verified by 14 fact-checking agencies from a wide range of sources, such as messaging apps, social platforms, and traditional news outlets. As mentioned above, these agencies are categorized into three groups (see Table~\ref{tab:source}. The websites for the 14 fact-checking agencies are detailed in Appendix \ref{app:url}.
The first group includes five Chinese fact-checking agencies identified as active by Duke Reporters\footnote{\url{https://reporterslab.org/fact-checking/}}, along with four other Chinese fact-checking agencies manually verified by experts as active and authoritative.
The second group corresponds to four English fact-checking agencies, including Politifact, Gossipcop, BS Detector, and FakeNewsCorpus. This group contains the Chinese counterparts of the English news fact-checked by the aforementioned agencies, which are collected through a carefully designed cross-lingual identical news retrieval method. An in-depth description of the method can be found in Sec.~\ref{sec:newimi}.
Group 3 exclusively covers the Weibo Community Management Center, from which news data was directly sourced from \textit{Weibo-21} dataset~\cite{WEIBO21}.

\begin{table}[h]
\caption{An overview of fact-checking agencies.}
\label{tab:source}
\begin{tabular}{cccc}
\hline
\textbf{Group} & \textbf{Fact-checking Agencies}\\
\hline
\multirow{9}{*}{Group 1} 
 & China Internet Joint Rumor Refuting Platform &\\
 & Tencent Jiaozhen & \\
  & China Daily Factcheck &\\
 & Taiwan FactCheck Center & \\
  & MyGoPen &\\
 & HKBU Factcheck & \\
  & HKU Annie Lab &\\
 & AFP Fact Check Asia & \\
  & Factcheck Lab &\\
\hline

\multirow{4}{*}{Group 2}
 & Politifact & \\
  & Gossipcop &\\
  & BS Detector&\\
  & FakeNewsCorpus &\\
 \hline
  Group 3 & Weibo Community Management Center  \\ 
 \hline
\end{tabular}
\end{table}
The overall \texttt{MCFEND} dataset contains 23,789 news pieces, including 7,959 sourced from the nine fact-checking agencies in Group 1, 6,702 related to the four English fact-checking agencies in Group 2, and 9,128 obtained from the Weibo Community Management Center in Group 3.
Similar to existing dataset construction~\cite{FAKENEWSNET, WEIBO21}, we collected the following information for each piece of news in any group:
(1) Multi-modal news content, including text, images, and metadata, e.g., timestamps;
(2) Multi-modal social context, including posts, comments, emojis, user profiles, and other metadata, e.g., like counts of comments. Table~\ref{tab:data} presents the detailed statistics of the \texttt{MCFEND} dataset.

\begin{table}[h]
\caption{Statistics of the \texttt{MCFEND} dataset.}
\label{tab:data}
\begin{tabular}{ccccc}
\hline
\textbf{Statistics} & \textbf{Group 1} & \textbf{Group 2} &\textbf{Group 3} & \textbf{Overall} \\
\hline
\#Total& 7,959 &6,702 &9,128 & 23,789\\  %23,974
\#Fake& 7,486 & 5,741 & 4,488 &17,715\\
\#Real& 473 & 9,61 & 4,640 &6,074\\
\hline
\#user & 235,215 &156,862  &458,800 &803,779 \\ 
\#posts & 58,299 & 41,600 & 70,814&170,713\\ 
\#comments & 262,342 & 328,465 & 1,512,095&2,102,902\\ 
\hline
\multirow{3}{*}{Timeframe} 
 & Mar. 2015 & Jan. 2015 & \camera{Dec. 2014} &Jan. 2015\\
 & - & - & - &-\\
 & Mar. 2023 & Mar. 2023 & Mar. 2021  &Mar. 2023\\
\hline
\end{tabular}
\end{table} 

\subsection{Dataset Construction}
\begin{figure*}[h]
 \centering
 \includegraphics[width=0.85\linewidth]{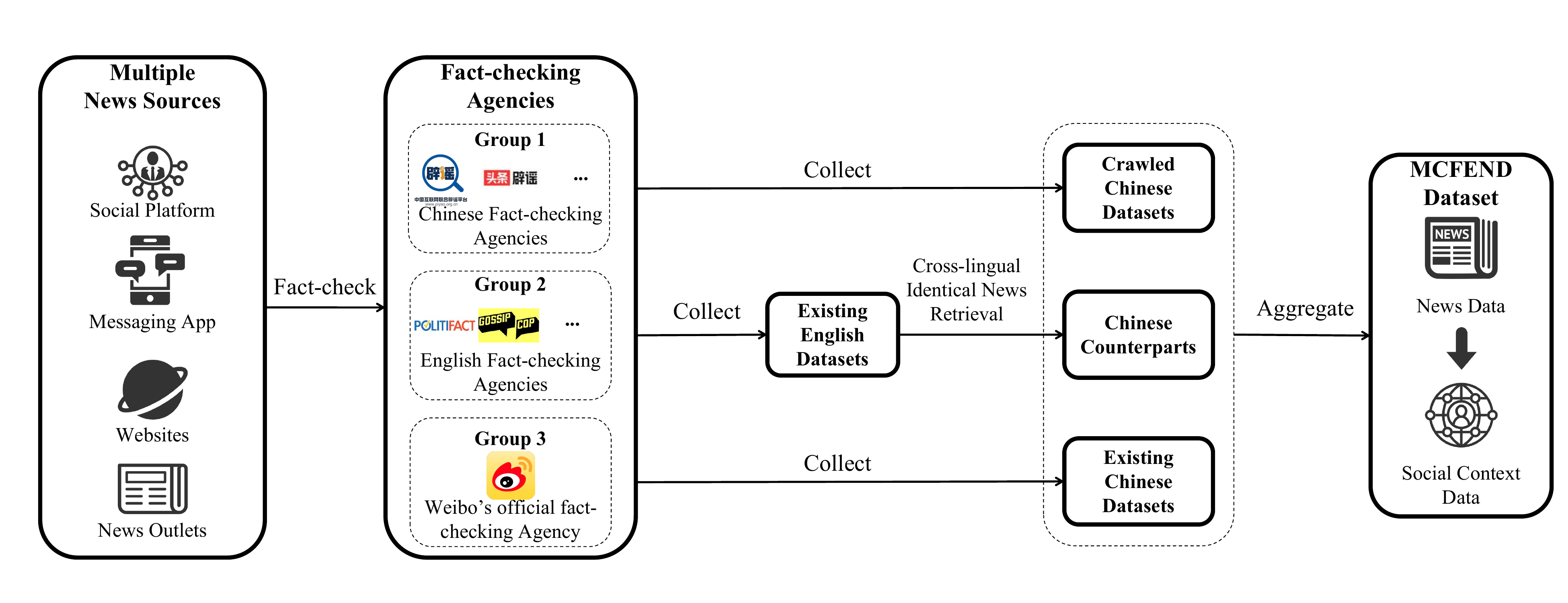}
 \caption{The process for constructing the \texttt{MCFEND} dataset.}
 \label{fig:construction}
\end{figure*}

In this subsection, we present the process of constructing the dataset for each group of the fact-checking agencies. Fig.~\ref{fig:construction} illustrates the entire process for the dataset construction.

\subsubsection{Group 1: Fact-checking Agencies Data Crawling}
\label{sec:factcheck}
Fact-checking agencies serve as a common source for labeling fake news detection datasets~\cite{support4, survey}. These agencies are typically operated by government entities, companies, or non-profit organizations, and they employ authoritative experts to assess the authenticity of news pieces originating from diverse sources, such as social platforms, messaging apps, and traditional online news outlets. As discussed in Sec.~\ref{sec:intro}, including a wider range of fact-checking agencies enhance the diversity of news sources in our dataset.

To maximize our coverage of news sources, we conducted web crawling to collect data from all active Chinese fact-checking agencies, encompassing the five Chinese fact-checking agencies identified as active by Duke Reporters, in addition to four other Chinese fact-checking agencies that were manually verified as active and authoritative. In the case where the labels on some fact-checking agencies, e.g., AFP Fact Check Asia and Factcheck Lab, are presented in the form of images, we utilized an optical character recognition method called Tesseract-OCR to retrieve such labels.\footnote{\url{https://github.com/tesseract-ocr/tesseract}}

\subsubsection{Group 2: Cross-lingual Identical News Retrieval}
\label{sec:newimi}
To further diversify our news sources, we employ a cross-lingual identical news retrieval method to obtain the corresponding Chinese equivalents of both real and fake English news within several well-known datasets, including \textit{FakeNewsNet}~\cite{FAKENEWSNET}, \textit{KaggleFakeNews}~\cite{kagglefakenews}, and \textit{FakeNewsCorpus}~\cite{support3}. \textit{FakeNewsNet} consists of two detailed sub-collections sourced from distinct fact-checking organizations, specifically Politifact and Gossipcop. \textit{KaggleFakeNews} contains news gathered from 244 sources classified as ``unreliable or otherwise questionable'' by the BS Detector, a browser extension that assesses the reliability of websites by comparing them to a professionally curated list.
\textit{FakeNewsCorpus} is a dataset consisting of news related to the 2016 US elections. The news pieces in this dataset are manually annotated by its authors. 

We consider the BS Detector and the authors of the \textit{FakeNewsCorpus} as two distinct fact-checking agencies. By incorporating these three datasets, we effectively introduce data from four additional fact-checking agencies, enabling us to collect news from a wider range of English news sources.

For each news piece in these datasets, we executed the following steps to identify its corresponding Chinese counterpart:
\begin{itemize}
    \item Step 1: Translation. We utilized the Baidu Translation API to translate the headlines of the English news into Chinese.\footnote{\url{http://fanyi-api.baidu.com/}}
    \item Step 2: Chinese News Retrieval with Google News.\footnote{\url{https://news.google.com/}} Google News provides extensive and up-to-date news coverage from sources worldwide. We configured the language and region of interest as ``Chinese (China)'' and employed the translated Chinese news headline as the search query. Search engines typically sort results by relevance. We assumed that the top five returned news pieces were the most relevant Chinese counterparts to the original English news. Subsequently, we crawled the top five returned news pieces.
    \item Step 3: Cross-lingual News Similarity Calculation. To determine the degree of similarity between the Chinese news retrieved in the previous step and the original English news, \camera{we employed the state-of-the-art cross-lingual news similarity calculation system~\cite{newsimimodel}, which ranked 1st in the SemEval2022 Task 8 challenge~\cite{SemEval-2022-Task-8} with a Pearson correlation coefficient of 0.818 on the official evaluation set.} Specifically, we calculated the similarity score between the retrieved Chinese news and the original English news. The Chinese news with the highest similarity score was preserved in our \texttt{MCFEND} dataset. The strong performance of the system effectively ensures the consistency of misleading content across different languages.
    \item Step 4: Label Assignment. The authenticity label of the original English news is used to label its Chinese counterpart, that is, the Chinese news with the highest similarity score.
\end{itemize}
The method inherently retains human-written news content. In contrast to directly utilizing Chinese news content generated by machine translator, our approach avoids unnatural textual expressions that could potentially introduce noise to the models.

\subsubsection{Group 3: Weibo News Collection}
Group 3 consists solely of news sourced from the Weibo. For this group, we directly utilized news data in the \textit{Weibo-21} dataset~\cite{WEIBO21}, the largest Chinese fake news detection dataset on Weibo.

\subsubsection{Social Context Collection}
Relying solely on news content may be inadequate for detecting fake news, as fake news content is often meticulously crafted to deceive the public. Social platforms offer an invaluable source of supplementary information in the form of social context features~\cite{FAKENEWSNET, survey, Tree-LSTM, Tree-RvNN, defend,Tree-Transformer,BERT-EMO}, capturing user interactions and social behaviors within the social platform environment. Thus, to incorporate such important features, we collected social context data, such as posts, comments, user profiles, etc., on the largest social platform in China, Weibo.\footnote{While Weibo serves as the social context source for all collected news pieces, it also acts as an independent news source.}

The process of collecting social context aligns closely with the approach detailed in~\cite{FAKENEWSNET} for gathering social context from Twitter. Firstly, for news pieces that have headlines, we created search queries for associated posts on Weibo using the headlines. For news pieces without headlines, we utilized the Jieba tool to tokenize the textual content of the news and extract the top five keywords, which were then used as search queries.\footnote{\url{https://github.com/fxsjy/jieba}} During this process, we removed special characters from the search queries to eliminate unnecessary noise. All matching posts were considered relevant and included. We then retrieved user responses to these posts, including comments, reposts, and likes. Additionally, upon identifying all users involved in the news propagation process, we collected metadata for these users, such as their usernames and profiles. As shown in Table~\ref{tab:data}, we assembled a comprehensive set of relevant social context data, which includes 170,713 posts and 2,102,902 comments from 803,779 distinct users.\footnote{Individual users may engage in the social context of news collected from fact-checking agencies across different groups.}
\subsubsection{Post-collection Processing}
After collecting all news pieces and their corresponding social context, we conducted three post-collection processing steps:
\begin{itemize}
    \item Step 1: Text Cleaning. To enhance data quality and eliminate unnecessary noise, we conducted text cleaning on text within both news content and social context. This cleaning process involved removing HTML tags, punctuation, white spaces, stop words, and prefix headings. 
    \item Step 2: Deduplication. The raw data contained multiple duplications. As a result, we removed redundant news and social context data to avoid unnecessary repetitions.
    \item Step 3: Label Mapping. Different fact-checking agencies employ diverse fine-grained labels to express degrees of authenticity (e.g., true, mostly true, and inconclusive). To ensure consistency, we designed a label mapping strategy to standardize the original labels. 
    \camera{Please refer to Appendix~\ref{app:label_mapping} for details of our label mapping strategy.}
\end{itemize}

\subsection{Comparison of the Three Groups}
Our \texttt{MCFEND} dataset gathers news from 14 fact-checking agencies, categorized into three distinct groups. As previously noted, fake news originating from diverse sources differs in both content and social context~\cite{support1,support2,support3, support4,support5}. To quantify these differences between the three groups, we conducted analyses on these two aspects, which are considered significant clues for fake news detection~\cite{survey, support4}.

\begin{figure}[htbp]
    \centering
    \subfloat[Textual feature.]{\includegraphics[width=.49\columnwidth]{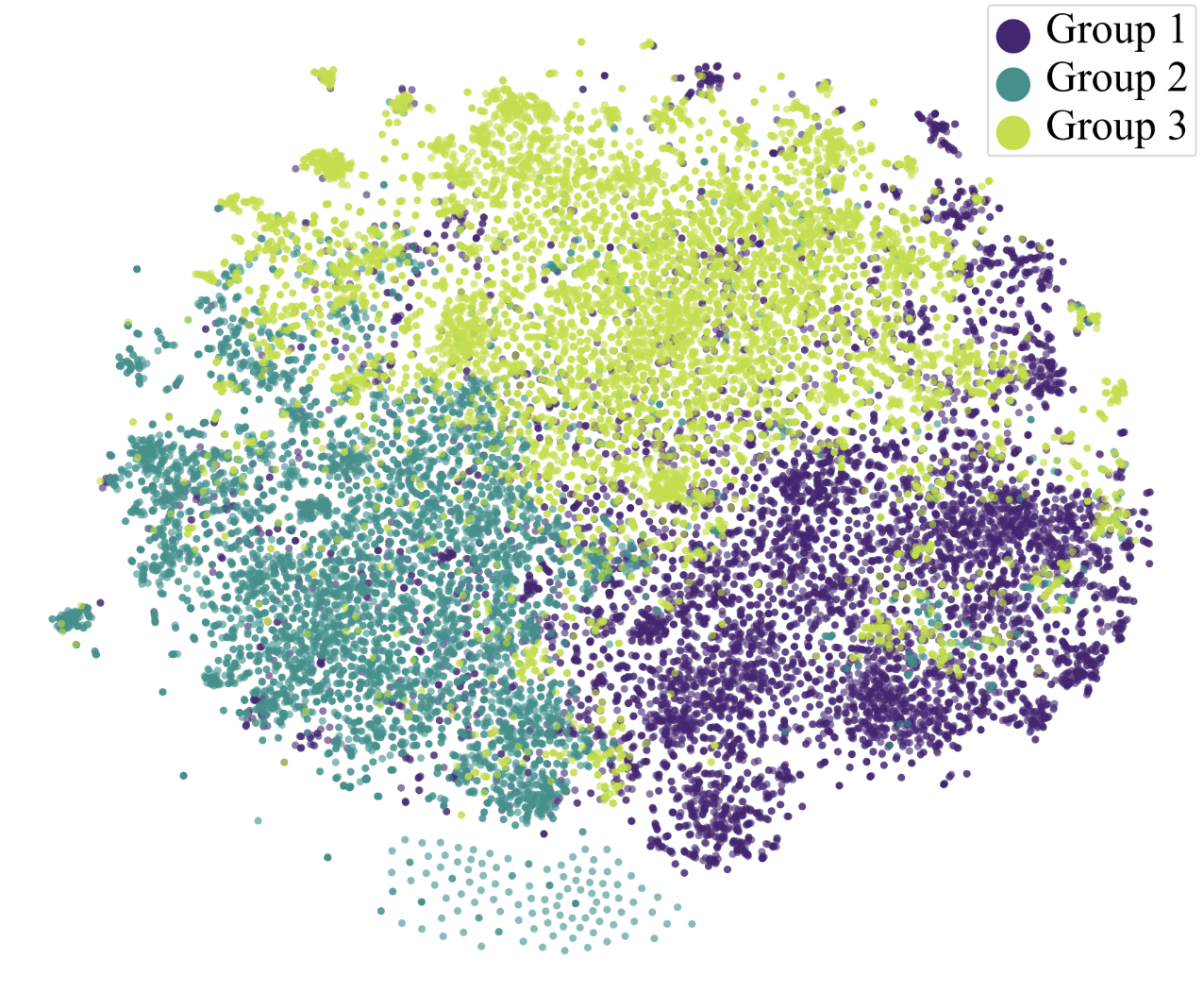}} \hspace{2pt}
    \subfloat[Social emotion feature.]{\includegraphics[width=.49\columnwidth]{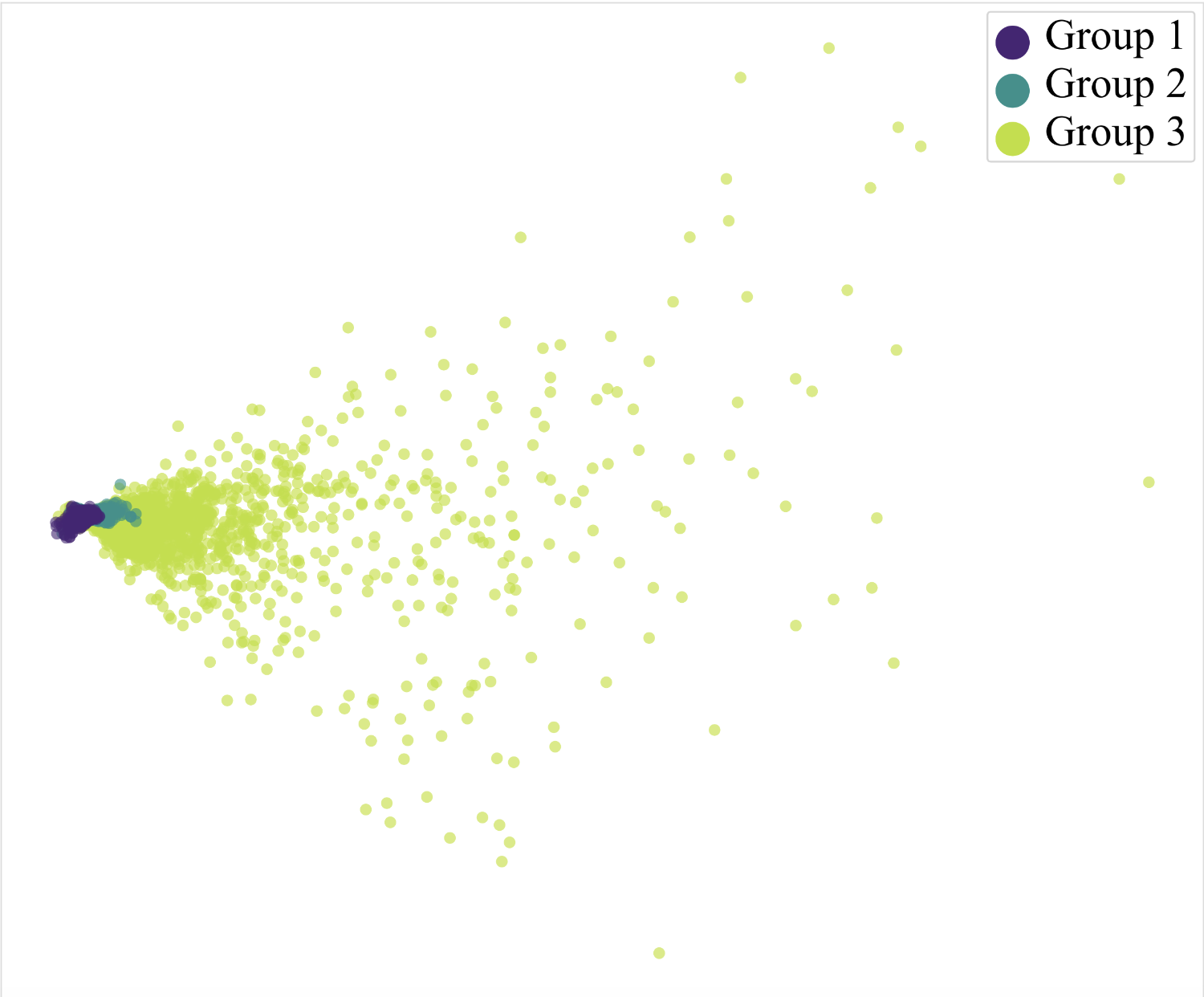}}
    \caption{Visualization of textual and social emotion features for news collected from three distinct groups of fact-checking agencies.}
    \label{fig:combined}
\end{figure}

Regarding content analysis, we employed a pretrained Chinese Sentence-BERT model~\cite{sentence-bert} to generate text representations for each news piece. Consequently, each news piece was represented as a 768-dimensional vector based on its textual information. This representation serves as an encompassing approximation of the textual features of the news. To visually depict the differences, we applied t-SNE~\cite{tsne} to reduce the dimensionality of these vectors to two.\footnote{\camera{Other dimensionality reduction techniques like PCA and UMAP are applicable, but due to space limits, we only present the results of t-SNE here.}} As shown in Fig.~\ref{fig:combined} (a), the textual features of news pieces collected from fact-checking agencies are distinct. This result validates the significant difference in their textual characteristics.

For the social context, we considered the significant social emotion features~\cite{BERT-EMO}, which is a 275-dimensional vector representing the emotions evoked within the social context surrounding the news pieces and has been verified to exhibit distinctions between fake and real news~\cite{BERT-EMO}. Similar to the analysis above, we reduced the dimensionality of the social emotion feature to two. As illustrated in Fig.~\ref{fig:combined} (b), the social emotion features associated with news pieces from Weibo demonstrate a more scattered pattern, whereas those from other sources are clustered. These results illustrate the substantial differences in their social context characteristics.

In summary, our analysis of both content and social context provides strong evidence of differences between the three groups. We also consider other factors like text length and dual emotions between the news publisher and associated users in the social context. Due to space limitations, we omit the analysis here.

\section{Experiments}\label{sec:experiment}
We conducted experiments to evaluate the performance of %existing 
representative fake news detection methods on our newly proposed \texttt{MCFEND} dataset. Specifically, we aim to answer the following evaluation questions (EQs):

\begin{itemize}
    \item[EQ1]: Are existing methods, which have demonstrated effectiveness on the existing Weibo datasets, capable of maintaining their performance when being applied to news collected from different sources?
    \item[EQ2]: Can training with multi-source data enhance the robustness of existing methods in detecting fake news in real-world scenarios, which involve multiple sources?
    \item[EQ3]: Can training with multi-source data enhance the robustness of existing methods in detecting fake news originating from previously unseen news sources?
\end{itemize}
\camera{The detailed experimental setups are provided in Appendix~\ref{app:exp}.}

\subsection{Baselines}
To address the above EQs, we carefully selected six baseline models spanning the two widely recognized categories of fake news detection approaches: content-based and social context-based methods~\cite{survey, MR2, support5, faknow}, and constructed the baseline benchmarks. Their implementation details are as follows.

\subsubsection{Content-based Methods}
Content-based methods rely on solely the textual or visual contents of the news. We adopted two representative types of content-based models, \textit{uni-modal models} and \textit{multi-modal models}.

\textit{Uni-modal models} focus on the textual content of the news. We used BERT~\cite{BERT} and RoBERTa~\cite{ROBERTA} as contextualized encoders to encode the textual content. 
Then, the representation of the ``[CLS]'' special token is used for prediction. The implementation of BERT and RoBERTa in this study is based on their respective Chinese base versions, i.e., BERT-base-Chinese\footnote{\url{https://huggingface.co/bert-base-chinese}} and RoBERTa-wwm-Base.\footnote{\url{https://huggingface.co/hfl/chinese-roberta-wwm-ext}}
\textit{Multi-modal models} encode both text and images in the input news. We consider two multi-modal baselines: CLIP~\cite{CLIP} and CAFE~\cite{CAFE}. 
CLIP~\cite{CLIP} is a pretrained model for images and text. We input the image and text of the news into CLIP to obtain a joint representation of the visual and textual content. 
This joint representation is then utilized for predictions. 
We implement CLIP based on its Chinese version.\footnote{\url{https://github.com/OFA-Sys/Chinese-CLIP}} CAFE~\cite{CAFE} is an ambiguity-aware fake news detection method. 
Specifically, it integrates uni-modal features produced by BERT \cite{BERT} for text and ResNet~\cite{Resnet} for images, along with cross-modal correlations. 
It relies on uni-modal features when cross-modal ambiguity is weak and relies on cross-modal correlations when cross-modal ambiguity is strong. 
CAFE demonstrates superior fake news detection performance on the \textit{Twitter}~\cite{Twitter} and \textit{Weibo-16}~\cite{WEIBO16} datasets, respectively. 
It stands as the state-of-the-art content-based approach to this task.

\subsubsection{Social Context-based Methods}
Social context-based methods are typically categorized into three groups: tree-based, modal fusion-based, and graph-based~\cite{MR2}. \camera{Since our \texttt{MCFEND} dataset contains news from diverse sources and lacks the necessary cross-source user/news interactions to build effective graphs and trees, we included only the \textit{modal fusion-based models}, excluding the graph-based and tree-based models.}

\textit{Modal fusion-based models} integrate information from both news content and social context. In our study, we considered two representative baseline models for this category: dEFEND~\cite{defend} and BERT-EMO~\cite{BERT-EMO}. 
The dEFEND model~\cite{defend} utilizes a sentence-comment co-attention sub-network to exploit news contents and comments in the social context to jointly capture explainable top-k check-worthy sentences and comments for fake news detection. 
On the other hand, BERT-EMO~\cite{BERT-EMO} enhances a BERT-based fake news detector by incorporating dual emotion features that represent both the emotions and the relationship between news and comments within the social context. 
Note that the BERT-EMO model has demonstrated outstanding performance in fake news detection, achieving the highest reported performance on the \textit{Weibo-20} dataset~\cite{BERT-EMO}. Besides, our preliminary experiments on the \textit{Weibo-21} dataset~\cite{WEIBO21} have also shown that BERT-EMO achieved an impressive F1-score of 0.943, outperforming all other methods. 
The results show that BERT-EMO as the state-of-the-art social context-based approach in the fake news detection task.

\subsection{Cross-source Evaluation}
To address EQ1, we performed cross-source evaluations on the baseline models. 
Specifically, we evaluated the performance of baseline models trained exclusively on Weibo data, focusing on their application to diverse news sources. 
Our findings, shown in Table~\ref{tab:cross}, reveal significant variations in performance across different test groups, underscoring the challenge of applying models trained on a single-source dataset to news originating from varied sources.
\begin{table}[h]
\centering
\caption{The performance of the baselines in cross-source evaluation. 
The training data for all baselines in cross-source evaluation is exclusively sourced from Weibo. 
The overall results are calculated using test data from all groups. The highest number in each group is in bold.}
\label{tab:cross}
\begin{tabular}{cccc}
\hline
\textbf{Model}                    & \textbf{Test Data} & \textbf{Accuracy}   & \textbf{Macro F1}   \\ \hline
\multirow{4}{*}{BERT}             & Group 1              &\textbf{0.830±0.018}&\textbf{0.521±0.009}\\
                                  & Group 2              &0.453±0.027&0.407±0.017\\
                                  & Group 3              &0.817±0.006&0.816±0.006\\
                                  & Overall              &0.719±0.008&0.673±0.007\\ \hline
\multirow{4}{*}{RoBERTa}          & Group 1              &0.124±0.008&0.123±0.007\\
                                  & Group 2              &\textbf{0.763±0.038}&\textbf{0.539±0.016}\\
                                  & Group 3              &0.883±0.006&0.883±0.006\\
                                  & Overall              &0.595±0.011&0.582±0.009\\ \hline
\multirow{4}{*}{CLIP}             & Group 1              &0.769±0.011&0.503±0.006\\
                                  & Group 2              &0.500±0.043&0.458±0.030\\
                                  & Group 3              &0.895±0.012&0.895±0.012\\
                                  & Overall              &0.741±0.009&0.711±0.009\\
                                  \hline
\multirow{4}{*}{CAFE}             & Group 1              &0.586±0.013&0.431±0.006\\
                                  & Group 2              &0.343±0.032&0.335±0.026\\
                                  & Group 3              &0.891±0.001&0.891±0.001\\
                                  & Overall              &0.635±0.014&0.621±0.013\\ \hline
\multirow{4}{*}{dEFEND}           & Group 1              &0.656±0.031&0.452±0.009\\
                                  & Group 2              &0.511±0.058&0.400±0.026 \\
                                  & Group 3              &0.743±0.016&0.743±0.016\\
                                  & Overall              &0.709±0.002&0.701±0.003\\ \hline
\multirow{4}{*}{BERT-EMO}         & Group 1              &0.572±0.043&0.405±0.013\\
                                  & Group 2              &0.343±0.174&0.287±0.104\\
                                  & Group 3              &\textbf{0.943±0.010}&\textbf{0.943±0.010}\\
                                  & Overall              &\textbf{0.821±0.008}&\textbf{0.818±0.008}\\ \hline 
\end{tabular}
\end{table}

\camera{
The BERT-EMO model delivered the highest performance, achieving an accuracy of 0.821 and a macro F1 score of 0.818. However, while it scored a high macro F1 of 0.943 on data from Weibo (Group 3), its effectiveness significantly decreased on data from Group 1 and Group 2, with scores dropping to 0.572 and 0.343, respectively. 
Note that the pattern of performance variance was consistent across all baseline models, including BERT, RoBERTa, CLIP, CAFE, and dEFEND, across different groups.
\emph{These findings provide a crucial insight in the responses to EQ1: Baseline models, even those considered state-of-the-art and trained on Chinese fake news detection datasets from Weibo, exhibit limited robustness when confronted with fake news from diverse sources in the wild.} 
One contributing factor to such performance decrease may be the significant difference in the content and social context feature, as shown in Fig.~\ref{fig:combined}, between news sourced from Weibo and news from other sources. 
This discrepancy also highlights a significant concern regarding the possible overestimation of the effectiveness of current Chinese fake news detection methods, underscoring the necessity for a thorough reevaluation before they are considered for practical application in real-world situations.}

\camera{A comparison in Table~\ref{tab:decrease} shows the average decrease in macro F1 score when moving from Weibo-sourced data to other sources, with a smaller decrease indicating greater source robustness. 
Our analysis found that models employing modal fusion approaches, integrating both text and social context, demonstrate stronger resilience against data source variability, with an average decrease of only 0.084. 
In particular, the modal fusion-based model, dEFEND, exhibited the greatest robustness, with a minimal decrease of 0.032. 
In contrast, uni-modal and multi-modal approaches saw larger decreases of 0.222 and 0.227, respectively. 
These results indicate that the model performance is greatly influenced by content pattern differences among sources, suggesting that exploring ways to mitigate these impacts is a valuable direction for future research.}

\begin{table}[h]
\caption{The comparison of the average performance decrease of all baseline categories, which represents the difference between overall performance and performance on data sourced from Weibo. A smaller performance decrease indicates greater cross-source performance robustness.}
\label{tab:decrease}
\begin{tabular}{cc}
\hline
\textbf{Baseline Category} & \textbf{Average Macro F1 Decrease} \\ \hline
\textit{Uni-modal}          & -0.222  \\
\textit{Multi-modal}        & -0.227       \\
\textit{Modal fusion-based} & -\textbf{0.084}            \\ \hline
Average & -0.107 \\ \hline
\end{tabular}
\end{table}

\subsection{Multi-source Evaluation}\label{sec:multisource}
To address EQ2, we performed multi-source evaluations on the selected baseline models. Specifically, these baseline systems were trained with the complete train set of the \texttt{MCFEND} dataset, including news from all sources covered in our dataset and their corresponding social contexts.

\begin{table}[h]
\caption{The performance of the baselines in multi-source evaluation. The training data for all baselines in multi-source evaluation contains news from all sources. The overall results are calculated using test data from all groups. The highest number in each group is in bold.}
\label{tab:multi}
\begin{tabular}{cccc}
\hline
\textbf{Model}                    & \textbf{Test Data} & \textbf{Accuracy}   & \textbf{Macro F1} \\ \hline
\multirow{4}{*}{BERT}             & Group 1              &\textbf{0.934±0.004}&0.535±0.015\\
                                  & Group 2              &0.858±0.007&0.614±0.052\\
                                  & Group 3              &0.803±0.015&0.801±0.015\\
                                  & Overall              &0.862±0.005&0.800±0.012\\ \hline
\multirow{4}{*}{RoBERTa}          & Group 1              &0.939±0.002&0.518±0.012\\
                                  & Group 2              &\textbf{0.860±0.001}&0.501±0.013\\
                                  & Group 3              &0.844±0.011&0.844±0.011\\
                                  & Overall              &0.880±0.004&0.827±0.008\\ \hline
\multirow{4}{*}{CLIP}             & Group 1              &0.8891±0.020&0.5213±0.018\\
                                  & Group 2              &0.690±0.073&0.565±0.044\\
                                  & Group 3              &0.8518±0.005&0.851±0.005\\
                                  & Overall              &0.8178±0.015&0.7685±0.014\\ \hline
\multirow{4}{*}{CAFE}             & Group 1              &0.926±0.008&\textbf{0.563±0.016}\\
                                  & Group 2              &0.857±0.010&\textbf{0.675±0.007}\\
                                  & Group 3              &0.875±0.008&0.875±0.008\\
                                  & Overall              &0.887±0.007&0.845±0.007\\ \hline
\multirow{4}{*}{dEFEND}           & Group 1              &0.915±0.005&0.521±0.025\\
                                  & Group 2              &0.755±0.067&0.482±0.065\\
                                  & Group 3              &0.757±0.004&0.756±0.004\\
                                  & Overall              &0.791±0.007&0.774±0.007\\ \hline
\multirow{4}{*}{BERT-EMO}         & Group 1              &0.902±0.004&0.514±0.040\\
                                  & Group 2              &0.827±0.027&0.545±0.052\\
                                  & Group 3              &\textbf{0.937±0.006}&\textbf{0.937±0.006}\\
                                  & Overall              &\textbf{0.922±0.005}&\textbf{0.916±0.005}\\ \hline
\end{tabular}
\end{table}

The multi-source evaluation results, depicted in Table~\ref{tab:multi}, provide crucial insights. 
A comparison with the cross-source evaluation data (refer to Table~\ref{tab:cross}) reveals significant performance improvements in all baseline models upon integrating multi-source data for training. 
Notably, RoBERTa and CAFE show the most substantial gains, with their macro F1 scores increasing by 0.245 and 0.224, respectively. 
Such boost in performance can be attributed to the diverse range of fake news features presented by multi-source data, which enhances the models' ability to discern between fake and real news from different sources, helping to prevent models from overfitting to the specific characteristics of data from a single source like Weibo.
To offer a qualitative analysis of the enhancement from using multi-source data in the training process, we take the CAFE model as an example. 
When trained with Weibo data exclusively, the CAFE model fails to correctly identify the news pieces (c) and (d) shown in Fig.~\ref{fig:example} as fake. 
However, when trained with data from all the diverse sources encompassed in the \texttt{MCFEND} train set, the CAFE model exhibits the ability to accurately detect all of the presented fake news in Fig.~\ref{fig:example}. 

\emph{These findings address EQ2 by demonstrating that using multi-source data to train fake news detection models significantly improves their performance and robustness in real-world applications.} \texttt{MCFEND} dataset could be an invaluable asset for enhancing the detection of Chinese fake news across a variety of sources.

Additionally, the overall performance of almost all baseline models, except dEFEND, when they were trained and tested on multi-source data is lower than their performance when they were trained and tested on Weibo data exclusively. 
This finding addresses the challenge of developing algorithms capable of effectively distinguishing generic fake news features across news from various sources in real-world scenarios. 
\camera{Compared among baseline models, the state-of-the-art content-based and social context-based models, CAFE and BERT-EMO, stand out with macro F1 scores of 0.845 and 0.916, respectively. 
This implies that their relatively lower performance in cross-source evaluation suggests limitations in current training datasets. 
Thus, our diverse and comprehensive \texttt{MCFEND} dataset, designed to address these limitations, is crucial for advancing Chinese fake news detection in the wild.}

\section{Unseen Source Evaluation}
To address EQ3, we conducted an unseen source evaluation to assess the robustness of existing methods in detecting fake news from previously unencountered news sources. 
This involved training two versions of the BERT-EMO model, which had shown superior performance in both cross-source and multi-source evaluations, on distinct dataset compositions. 
Model A was trained exclusively on data from Group 3 (Weibo), while Model B incorporated data from both Group 1 (various Chinese news sources verified by fact-checking agencies) and Group 3 (Weibo).

\begin{table}[ht]
\centering
\caption{Macro F1 Score Comparison on Unseen Sources.}
\label{tab:results_comparison}
\begin{tabular}{lccc}
\hline
\textbf{Model} & \textbf{Train Data} & \textbf{Accuracy} & \textbf{Macro F1 Score} \\
\hline
Model A & Group 3 & 0.343±0.174 & 0.287±0.104\\
Model B & Groups 1 and 3 & \textbf{0.602±0.204}& \textbf{0.432±0.088}\\
\hline
\end{tabular}
\end{table}

We then evaluated both models using data from Group 2 (English news sources), which was new to each model. 
The results, detailed in Table~\ref{tab:results_comparison}, reveal that Model B, trained on a more diverse dataset, achieved a higher macro F1 score of 0.432, compared to Model A's score of 0.287. 
This finding addresses EQ3 by suggesting that leveraging multi-source data enhances the robustness of methods for detecting fake news from unseen sources. 
Therefore, when a new news source emerges in the future, but there is no available data from that platform for model training, a model trained on multi-source data can be expected to more accurately detect fake news from this unseen new source.

\section{Conclusion}\label{sec:conclusion}
In this work, we introduced the first multi-source benchmark dataset for Chinese fake news detection, termed \texttt{MCFEND}. Unlike existing Chinese fake news detection datasets that are based on a single news source, i.e., Weibo, \texttt{MCFEND} is constructed on (real and fake) news from multiple sources that were fact-checked by 14 authoritative fact-checking agencies. \haorui{To test the applicability of existing methods,} we conducted a systematic evaluation of six representative fake news detection models, including the state-of-the-art ones, in both cross-source and multi-source scenarios. Our experimental results reveal that models trained exclusively on Weibo data can hardly be applicable in real-world scenarios, where fake news typically originates from diverse sources. We also found that incorporating multi-source data into model training enhances the robustness of existing fake news detection methods. Our proposed \texttt{MCFEND} aims to be a benchmark dataset for Chinese fake news detection in the wild, which advances new effective methods in this research field. 
% One interesting future direction is to explore multi-source data conflicts in the context of Chinese fake news detection. Certain data repairing methods, e.g., AutoRepair~\cite{autorepair}, could be leveraged when designing effective algorithms for detecting Chinese fake news.

Our world is increasingly suffering from unrestrained spread of misinformation in many areas. We hope MCFEND can put forward research results that can combat misinformation and help make our world a better one.

\section{Acknowledgement}
\camera{This work was supported by the National Natural Science Foundation of China (No. 62202402), the Guangdong Basic and Applied Basic Research Foundation (No. 2022A1515011583 and No. 2023A1515011562), the Hong Kong RGC Early Career Scheme (No. 22202423), the Germany/Hong Kong Joint Research Scheme sponsored by the Research Grants Council of Hong Kong and the German Academic Exchange Service of Germany (No. G-HKBU203/22), the One-off Tier 2 Start-up Grant (2020/2021) of Hong Kong Baptist University (Ref. RC-OFSGT2/20-21/COMM/002), and the Startup Grant (Tier 1) for New Academics AY2020/21 of Hong Kong Baptist University.} 

\camera{We thank Ms.~Zihang Shan for her assistance in developing the cross-lingual identical news retrieval model and preparing the data. We also thank Ms.~Zixin Tang for her assistance in developing the CLIP baseline and executing the associated experiments.}

\bibliographystyle{miscellaneous/ACM-Reference-Format}
% \balance
\bibliography{reference}

% \newpage
\appendix

\section{The Fact-checking Agencies}
\label{app:url}
Table~\ref{tab:url} lists the websites for the 14 fact-checking agencies.

\begin{table}[b]
\centering
\caption{Fact-Checking Agencies.}
\label{tab:url}
\begin{tabular}{|p{0.45\linewidth}|p{0.45\linewidth}|}
\hline
\textbf{Fact-Checking Agency} & \textbf{URL} \\
\hline
China Internet Joint Rumor Refuting Platform & \url{https://www.piyao.org.cn/} \\\hline
Tencent Jiaozhen & \url{https://vp.fact.qq.com/} \\\hline
China Daily Factcheck & \url{https://www.chinadaily.com.cn/china/factcheck/} \\\hline
Taiwan FactCheck Center & \url{https://tfc-taiwan.org.tw/} \\\hline
MyGoPen & \url{https://www.mygopen.com/} \\\hline
HKBU Factcheck & \url{https://factcheck.hkbu.edu.hk/home/} \\\hline
HKU Annie Lab & \url{https://annielab.org/} \\\hline
AFP Fact Check Asia & \url{https://factcheck.afp.com/afp-asia/} \\\hline
Factcheck Lab & \url{https://www.factchecklab.org/} \\\hline
Politifact & \url{https://www.politifact.com/} \\\hline
Gossipcop & \url{https://www.gossipcop.com/} \\\hline
BS Detector & \url{https://github.com/selfagency/bs-detector} \\\hline
FakeNewsCorpus & \url{https://github.com/architapathak/FakeNewsCorpus} \\\hline
Weibo Community Management Center & \url{https://service.account.weibo.com} \\
\hline
\end{tabular}
\end{table}

\section{Label Mapping Strategy}
\label{app:label_mapping}
Various fact-checking agencies use distinct labels to indicate the truthfulness of information, such as ``rumor'', ``mistake'', and ``misleading''. 
To unify the label space in our dataset, we used a mapping strategy to standardize the original labels, as illustrated in Table~\ref{tab:map}. 
We also provide the English translations of these labels.

\section{Experimental Setup}
\label{app:exp}
Consistent with prior studies on detecting fake news, our evaluation framework utilizes accuracy and the macro F1 score as metrics. 
We divided the \texttt{MCFEND} dataset into training, validation, and test segments following a 7:1.5:1.5 ratio through random division to ensure our experimental results' reliability. 
This division was conducted three times, with each cycle running for 100 epochs on a single NVIDIA V100 GPU. We report the mean F1 and accuracy scores along with their standard deviations from the three iterations. 
All models were trained using the Adam optimizer with a batch size of 32. 
To increase training efficiency, we applied an early stopping rule, halting the training if no performance improvement was seen after 500 consecutive batches, ensuring model convergence.

Hyper-parameter selection was based on the validation set's performance. 
A comprehensive summary of the hyper-parameter settings for each model is provided in Table~\ref{tab:hyper}. 
For parameters not covered here, we suggest consulting the original publications for more details. 
We standardized the preprocessing of text to a maximum of 256 tokens and images to a uniform size of 224x224 pixels through center-cropping.
In scenarios where news items did not have associated images, we used a standard blank white image for training multi-modal models.

\begin{table}[b]
\centering
\caption{The strategy for mapping the authenticity labels by the fact-checking agencies to our standardized labels.}
\label{tab:map}
\begin{tabular}{c}
\includegraphics[width=\linewidth]{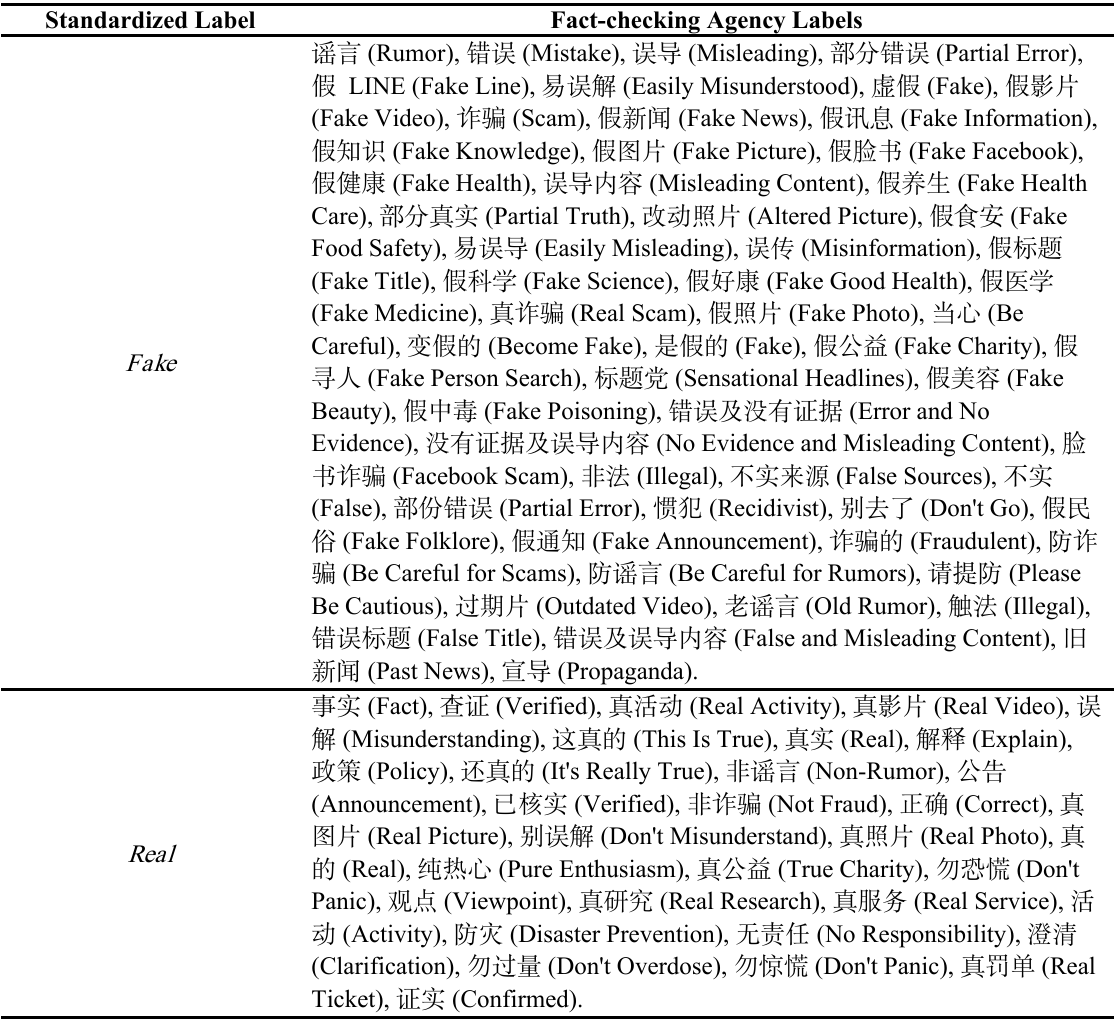}
\end{tabular}
\end{table}

\begin{table}[b]
\caption{Model-wise core hyper-parameter settings adopted in our implementations of the baselines.}
\label{tab:hyper}
\centering
\begin{tabular}{|c|C{6cm}|}
\hline
\textbf{Model} & \textbf{Hyper-parameters} \\
\hline
BERT & learning rate: 2e-5;\newline hidden size: 768; number of layers: 12;\newline number of attention heads: 12. \\
\hline
RoBERTa & learning rate: 2e-5;\newline hidden size: 768; number of layers: 12;\newline number of attention heads: 12. \\
\hline
CLIP & learning rate: 2e-5; project\_dim: 512.\\
\hline
CAFE & learning rate: 1e-3;\newline text\_embedding: fixed BERT embedding;\newline image\_embdding: fixed ResNet embedding. \\
\hline
dEFEND & learning rate: 2e-5;\newline text\_embedding: fixed BERT embedding;\newline project\_dim: 200;  number of RNN layers: 2.\\
\hline
BERT-EMO & learning rate: 1e-3; \newline hidden\_size (bidirectional-GRU): 32; \newline hidden\_size (fully connected layers): 32.\\
\hline
\end{tabular}
\end{table}
\end{document}